\documentclass[10pt,twocolumn,letterpaper]{article}

\usepackage{cvpr}      

\usepackage{graphicx}
\usepackage{amsmath}
\usepackage{amssymb}
\usepackage{booktabs}
\usepackage{multirow}
\usepackage{comment}
\usepackage{balance}

\usepackage[pagebackref,breaklinks,colorlinks]{hyperref}

\usepackage[capitalize]{cleveref}
\crefname{section}{Sec.}{Secs.}
\Crefname{section}{Section}{Sections}
\Crefname{table}{Table}{Tables}
\crefname{table}{Tab.}{Tabs.}

\newcommand{\algname}{ICIIA\xspace}
\newcommand{\inaturalisto}{iNaturalist 2019\xspace}
\newcommand{\inaturalist}{\emph{\inaturalisto}\xspace}
\newcommand{\femnisto}{FEMNIST\xspace}
\newcommand{\femnist}{\emph{\femnisto}\xspace}
\newcommand{\celebao}{CelebA\xspace}
\newcommand{\celeba}{\emph{\celebao}\xspace}
\newcommand{\imageneto}{ImageNet-1K\xspace}
\newcommand{\imagenet}{\emph{\imageneto}\xspace}
\newcommand{\ucfo}{UCF101\xspace}
\newcommand{\ucf}{\emph{\ucfo}\xspace}

\makeatletter
\newcommand\footnoteref[1]{\protected@xdef\@thefnmark{\ref{#1}}\@footnotemark}
\makeatother

\def\cvprPaperID{8675} 
\def\confName{CVPR}
\def\confYear{2023}

\begin{document}


\title{
  One-Time Model Adaptation to Heterogeneous Clients: 
  \\
  An Intra-Client and Inter-Image Attention Design
}

\author{
  Yikai Yan$^1$
  \and
  Chaoyue Niu$^1$
  \and
  Fan Wu$^1$
  \and
  Qinya Li$^1$
  \and 
  Shaojie Tang$^2$
  \and 
  Chengfei Lyu$^3$
  \and 
  Guihai Chen$^1$
  \\
  $^1$Shanghai Jiao Tong University \quad
  $^2$University of Texas at Dallas \quad
  $^3$Alibaba Group
}

\maketitle

\begin{abstract}
The mainstream workflow of image recognition applications is first training one global model on the cloud for a wide range of classes and then serving numerous clients, each with heterogeneous images from a small subset of classes to be recognized. From the cloud-client discrepancies on the range of image classes, the recognition model is desired to have strong adaptiveness, intuitively by concentrating the focus on each individual client's local dynamic class subset, while incurring negligible overhead. In this work, we propose to plug a new intra-client and inter-image attention (ICIIA) module into existing backbone recognition models, requiring only one-time cloud-based training to be client-adaptive. In particular, given a target image from a certain client, ICIIA introduces multi-head self-attention to retrieve relevant images from the client's historical unlabeled images, thereby calibrating the focus and the recognition result. Further considering that ICIIA's overhead is dominated by linear projection, we propose partitioned linear projection with feature shuffling for replacement and allow increasing the number of partitions to dramatically improve efficiency without scarifying too much accuracy. We finally evaluate ICIIA using 3 different recognition tasks with 9 backbone models over 5 representative datasets. Extensive evaluation results demonstrate the effectiveness and efficiency of ICIIA. Specifically, for ImageNet-1K with the backbone models of MobileNetV3-L and Swin-B, ICIIA can improve the testing accuracy to 83.37\% (+8.11\%) and 88.86\% (+5.28\%), while adding only 1.62\% and 0.02\% of FLOPs, respectively. Source code is available in the supplementary materials.
\end{abstract}

\vspace{-1em}
\section{Introduction}
\label{sec:intro} 




Nowadays, many vision models have been deployed to recognize client-side images, such as identifying everything with Google Lens, categorizing images in Google Photos, and taking photos to search for products on Amazon Shopping. Let's examine the workflow of Google Lens, a mobile camera app with strong image recognition capabilities, in detail. A global recognition model is trained  on the cloud using a large scale of labeled images, spanning a wide range of classes, including various species of animals and plants, documents, commodities, etc. Then, the recognition model will serve numerous app users once they take photos in daily life, and each user's photos tend to involve a few classes.


\begin{figure}[!t]
  \centering
  \includegraphics[width=0.95\columnwidth]{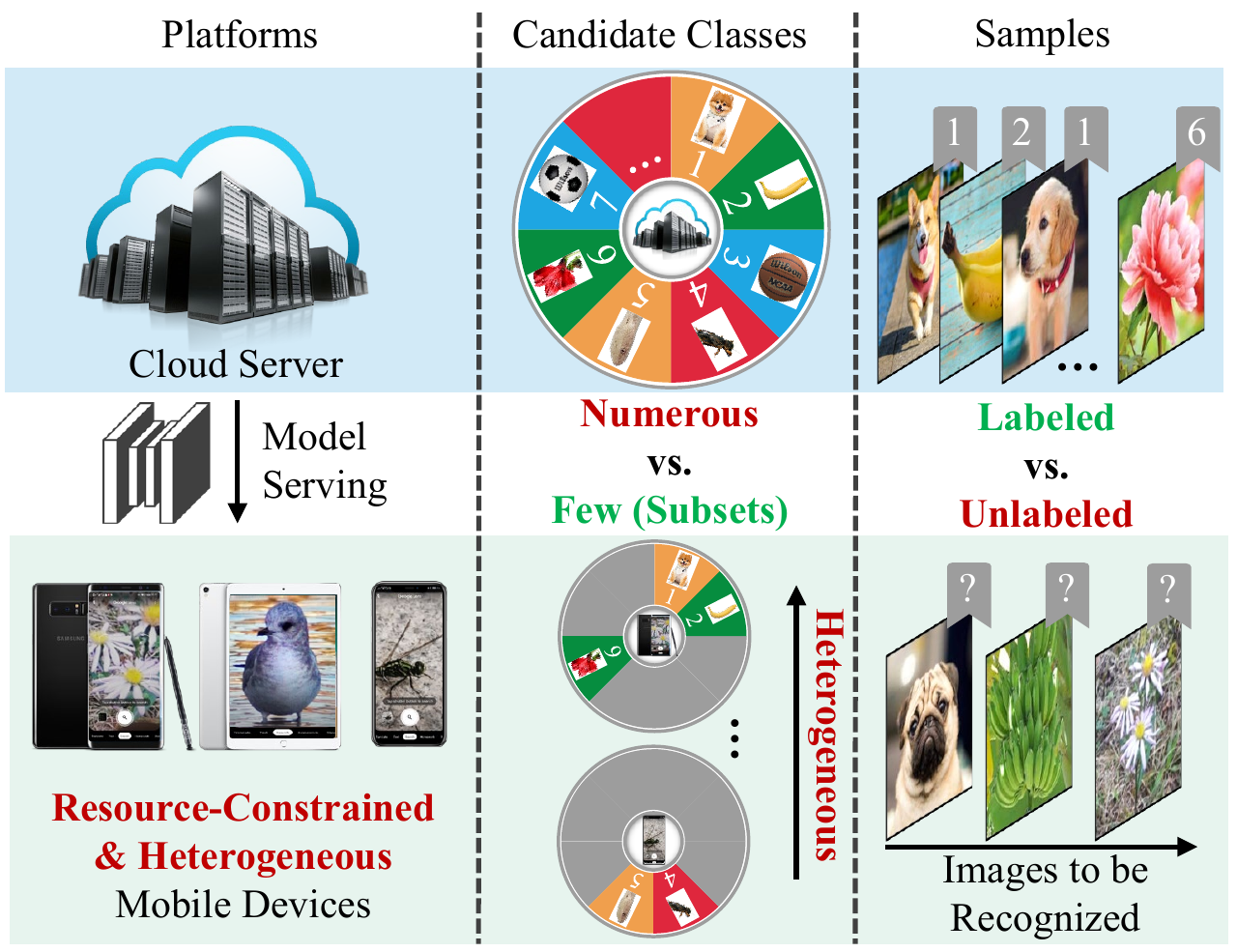}
  \caption{Image recognition from a new cloud-client view.}\label{fig:background}
  \vspace{-1em}
\end{figure}

By probing the image recognition scenario from a new cloud-client perspective, we find several important but often neglected characteristics, as illustrated in \cref{fig:background}. (1) The recognition model is optimized for the full set of classes on the cloud, whereas the images to be recognized on each individual client normally come from a small subset of classes. In addition, the class subsets of different clients also differ from each other. Moreover, as a client collects new images over time, the client's class subset will change dynamically. (2) Different from the cloud with labeled images for training, the images on each client are unlabeled for real-time recognition. It is also practically infeasible to let non-expert users label images and take their annotations as the ground truth. More generally, the missing of labels on the side of clients is an atypical setting of computer vision (CV) scenarios, compared with natural language processing (NLP) and recommendation applications, which typically predict user interactive behaviors (e.g., next input word, click or browse) and take practical user feedback as labels.  


The above cloud-client discrepancies raise the new requirement of strong adaptiveness on the image recognition model, from the cloud's global full set of classes to each client's local subset of classes, across different clients' heterogeneous class subsets, and over a certain client's dynamic image dataset. Intuitively, the focus of the recognition model was originally distributed on all the classes. When serving a specific client, if the focus can be concentrated on the client's class subset and further be dynamically adjusted as more unlabeled images are accumulated, the recognition accuracy can be significantly improved.             



To achieve model adaptation effectively and efficiently, there still exist several practical challenges. From model architecture, the design should be model-agnostic and lightweight, which is, however, non-trivial. In particular, the design should be compatible with existing backbone networks for image recognition (e.g., convolutional neural network (CNN), transformer, etc.), thereby inheriting their strong representation abilities and being convenient to be implemented and deployed. In addition, no matter whether the recognition model was originally deployed on the cloud to serve a large scale of clients or offloaded to resource-constrained and heterogeneous clients for local serving, the design would better introduce only a small size of parameters and negligible overhead to guarantee high efficiency (e.g., low response latency). Of course, in the resource-rich context, scaling up the design for better accuracy should be an available option. From learning algorithm, although client-specific fine-tuning is model-agnostic and even does not need extra parameters, the missing of labels and the resource constraints of clients make such type of methods inapplicable to image recognition. Therefore, it is highly desirable, yet challenging, to adapt the model only once on the cloud and circumvent on-client retraining.

In this work, we propose an \underline{I}ntra-\underline{C}lient and \underline{I}nter-\underline{I}mage \underline{A}ttention (\algname) module, which is plugable between the feature extractor and the classifier of an arbitrary backbone model for image recognition. After one-time training on the cloud, \algname can well adapt to heterogeneous clients. The key insight behind \algname's strong adaptiveness is that when serving a specific client, given a target image, \algname retrieves the relevant images from the client's historical images and enables the recognition model to concentrate on the local dynamic subset of classes, thereby calibrating the features and the recognition results. In contrast, the conventional mainstream mode is recognizing the images independently, failing to exploit each individual client's historical unlabeled images. The underlying building block of \algname is the celebrated multi-head self-attention mechanism, which was first proposed in NLP~\cite{nips/2017/vaswani/transformer} and later extended to computer vision~\cite{iclr/2021/dosovitskiy/vit}, and has achieved the state-of-art performance by capturing the intra-sample dependencies among words or image patches. At a totally different level, \algname intends to capture the inter-image dependencies for each individual client. In particular, the attention operation is performed over the separate pool of a certain client's local images to be recognized. Through naturally inputting the personalized images from the client rather than manually inserting additional client-specific parameters, \algname can achieve adaptiveness in a desired one-for-all way. Further, as the client accumulates more unlabeled images for attention, the performance of \algname will become better. Another key difference from the conventional multi-head self-attention mechanism is in linear projection, which originally needs a large size of parameters and incurs high overhead. \algname partitions the input features into several blocks and linearly projects for each block individually. To mitigate the side effect of isolating features in different partitions, \algname further shuffles the features across partitions. Through the partitioned linear projection with feature shuffling, \algname reduces the parameter size to the reciprocal of the number of partitions, while incurring only a slight drop in accuracy, compared with the conventional linear projection. By varying the number of partitions, \algname can well balance efficiency and accuracy.  




\begin{figure*}[!t]
  \centering
  \includegraphics[width=0.95\textwidth]{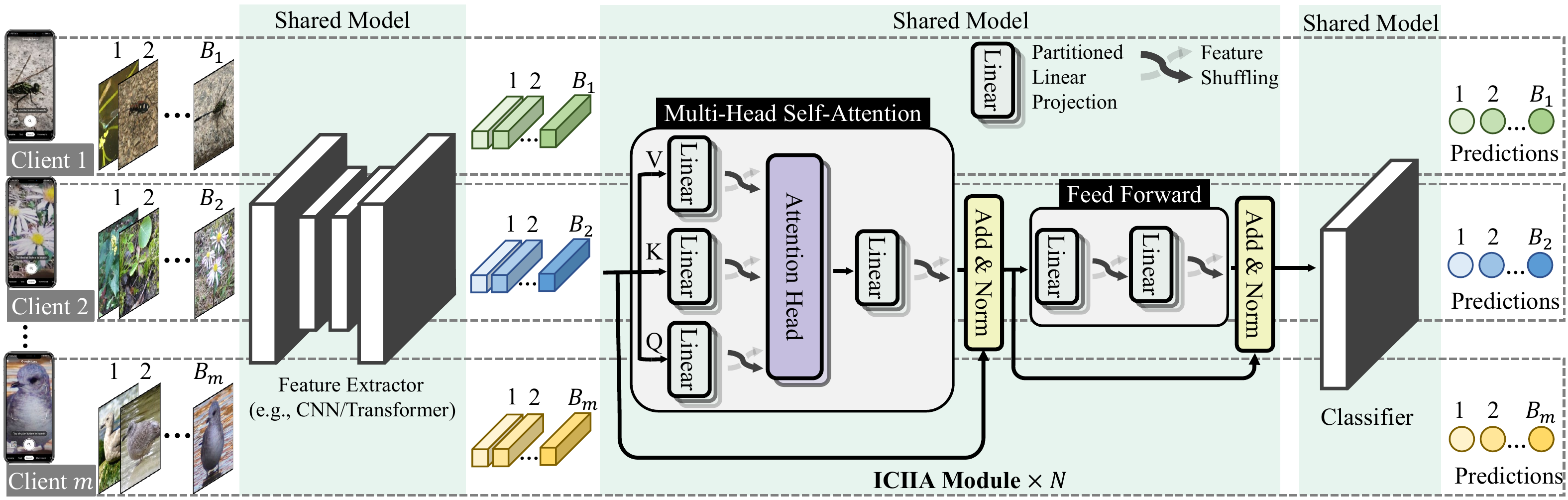}
  \caption{Overall model architecture. The new \algname module is plugable between the feature extractor and the classifier of an arbitrary image recognition backbone network for effective and efficient adaptation to heterogeneous clients.}\label{fig:overall}
  \vspace{-1em}
\end{figure*}

We summarize our key contributions as follows:
\begin{itemize}
  \item From a new cloud-client perspective to view the image recognition application, we find the discrepancies and the variations on the range of candidate classes. We further identify the new requirement of one-time model adaptation to all the heterogeneous clients under the practical setting of unlabeled images.
  \item We propose a plugable \algname module upon existing backbone image recognition models. To the best of our knowledge, \algname is the first to leverage the attention mechanism to mine intra-client and inter-image dependencies, thereby achieving strong model adaptiveness. 

  \item We further propose partitioned linear projection with feature shuffling to deal with the efficiency problem of the attention layer's conventional linear projection in serving a large scale of resource-constrained clients.
  \item We extensively evaluate \algname, with several baselines for comparison and variants for ablation study. Evaluation results reveal \algname's effective and efficient adaptation to heterogeneous clients, significant improvements over the original backbone models and tuning methods, and the necessity of each ingredient.
\end{itemize}

\section{Related Work}
\label{sec:related}


\paragraph{Model adaptation}

Adapting the global model trained on the cloud to serve heterogeneous clients is an important problem in practice. Rather than the cloud-to-client adaptation considered in this work, existing literature mainly considered task-to-task adaptation, either from upstream to downstream~\cite{naacl/devlin/2019/bert,icml/2019/houlsby/adapter,nips/2020/brown/gpt3,nips/2020/cai/tinytl,cvpr/2020/he/probe,acl/2021/mahabadi/hyper,acl/2021/li/prefix,emnlp/2021/lester/prompt,nips/2021/mahabadi/compacter,cvpr/2022/he/mae,acl/2022/zaken/bitfit,iclr/2022/hu/lora,eccv/2022/jia/vpt,corr/2022/lian/ssl} (e.g., from masked auto-encoding to classification), from one domain to another~\cite{icml/2018/li/explicit,cvpr/2019/guo/spottune,ijcv/2022/zhou/vpt,cvpr/2022/du/l2p,cvpr/2022/du/cocoop,cvpr/2022/wang/domain,cvpr/2022/li/domain,cvpr/2022/ye/night} (e.g., from natural images to synthesized images or from daytime scenes to nighttime scenes), or across multiple modalities~\cite{icml/2021/radford/clip,icml/2021/jia/align,cvpr/2022/lu/prompt,cvpr/2022/sung/vladapter,cvpr/2022/chen/visualgpt} (e.g., natural language and vision). A few recent work~\cite{aaai/2021/yu/meta,kdd/2022/yan/mpda,corr/2022/ding/label} considered cloud-to-client adaptation in e-commerce recommendation or next-word prediction with the local samples naturally labeled by each client, which differs from the label missing setting of our image recognition scenario. Among these existing work, one line resorted to fine-tuning the pre-trained model on the labeled data of each target task and elaborated on developing new pre-training techniques to facilitate the subsequent fine-tuning process~\cite{naacl/devlin/2019/bert,nips/2020/brown/gpt3,icml/2021/radford/clip,icml/2021/jia/align,cvpr/2022/he/mae} or designing new fine-tuning algorithms~\cite{icml/2018/li/explicit,cvpr/2019/guo/spottune,nips/2020/cai/tinytl,cvpr/2020/he/probe,acl/2022/zaken/bitfit,iclr/2022/hu/lora,corr/2022/lian/ssl,cvpr/2022/wang/domain}.
Other works further proposed to modify the backbone model, 
by inserting learnable task-specific adapters~\cite{icml/2019/houlsby/adapter,nips/2021/mahabadi/compacter,acl/2021/mahabadi/hyper,cvpr/2022/sung/vladapter,cvpr/2022/li/domain} 
or providing prompts (i.e., a small number of learnable tokens associated with each task) as additional inputs to the NLP models~\cite{acl/2021/li/prefix,emnlp/2021/lester/prompt} or recently to the CV models~\cite{eccv/2022/jia/vpt,ijcv/2022/zhou/vpt,cvpr/2022/lu/prompt,cvpr/2022/du/l2p,cvpr/2022/du/cocoop}. The latter method was also known as ``prompt tuning''. Compared with model adaptation to a few target tasks normally with labeled samples, the cloud-to-client recognition model adaptation studied in this work, however, involves a large number of target clients with only unlabeled images, making previous design inapplicable.



Different from cloud-based model training, an emerging federated learning~\cite{aistats/2017/mcmahan/fl} framework enables heterogeneous clients to jointly train a global model without uploading local data. Much related work also focused on the need of adapting the global model to each client, thereby mitigating data heterogeneity~\cite{corr/2020/mansour/three}. The designs mainly fall into the pattern of letting each client maintain a personalized model with client-specific parameters, by leverging
multi-task learning~\cite{nips/2017/smith/flmulti},
meta learning~\cite{corr/2018/chen/meta,corr/2019/jiang/flmaml,nips/2020/fallah/flmaml},
hypernetworks~\cite{cvpr/2022/ma/layer},
knowledge distillation~\cite{icml/2021/zhu/datafree,cvpr/2022/shen/pfed,cvpr/2022/zhang/flkd},
model splitting~\cite{corr/2020/liang/lg,nips/2021/luo/fear},
model pruning~\cite{corr/2020/li/lotteryfl},
or loss function regularization~\cite{nips/2020/dinh/envelope,icml/2021/li/ditto}.
However, federated learning normally require local training with labeled samples, which are unavailable in the scenario of image recognition.


\vspace{-1.2em}
\paragraph{Attention mechanism} For NLP and recommendation tasks, each sample is naturally represented by a sequence of words or user behaviors. The attention mechanism has been widely adopted to mine the dependencies within the sequence, such that the model can focus on the related words or behaviors for more accurate prediction~\cite{nips/2017/vaswani/transformer,kdd/2018/zhou/din,aaai/2019/zhou/dien}. In particular, Vaswani et al.~\cite{nips/2017/vaswani/transformer} proposed a novel architecture, called transformer, based on multi-head attention with residual connections~\cite{cvpr/2016/he/resnet} and layer normalization~\cite{corr/2016/ba/layernorm}, achieving state-of-the-art performance initially on NLP tasks and later on CV tasks~\cite{iclr/2021/dosovitskiy/vit, cvpr/2022/liu/swin}. Motivated by the sequential data format in NLP and recommendation, we regard the images to be recognized on each individual client as a sequence, rather than as separate images like in the mainstream recognition models. We thus introduce the attention mechanism and further deal with the efficiency issue of serving a large number of resource-constraint clients. 


\section{Design and Analysis}
\label{sec:approach}

\subsection{Overall Architecture}

As shown in \cref{fig:overall}, besides a feature extractor and a classifier in an arbitrary backbone recognition model, all the heterogeneous clients will share the new \algname module in our model design. Specifically, for each client, the images to be recognized are first input to the feature extractor. The extracted features are then fed into the \algname module. Given a target image, \algname essentially mines the intra-client and inter-image dependencies to retrieve the relevant historical images, and calibrates the features to let the classifier focus on the client’s local data distribution (\cref{sec:approach:attn}). Such a model architecture design can adapt to each heterogeneous client after one-time training on the cloud and dynamically improves the performance as the client accumulates more unlabeled images for attention. Furthermore, to reduce the overhead, the linear projections in the \algname module can be replaced with the proposed partitioned linear projection (\cref{sec:approach:trick1}) with feature shuffling (\cref{sec:approach:trick2}). The overhead analysis regarding the number of parameters and FLOPs is described in \cref{sec:approach:overhead}. Finally, the calibrated features from the \algname module are input into the classifier to generate the final recognition result.

\subsection{Intra-Client and Inter-Image Attention}
\label{sec:approach:attn}

The \algname module consists of a multi-head self-attention layer~\cite{nips/2017/vaswani/transformer} and a feed-forward network with residual connections and layer normalization. The multi-head self-attention layer takes the features of both the historical images and the target image from a certain client as the query vector (Q), the key vector (K), and the value vector (V), linearly projects each vector to $H$ representation subspaces, and computes the scaled dot-product attention scores with $H$ attention heads. Then, the features of the target image are calibrated using the weighted average of the features of the historical images, where the weights take the attention scores. The calibrated features are once again linearly projected and finally fed into the feed-forward network.


\begin{figure}[t]
  \centering
  \begin{subfigure}{0.96\columnwidth}
    \centering
    \includegraphics[width=\textwidth]{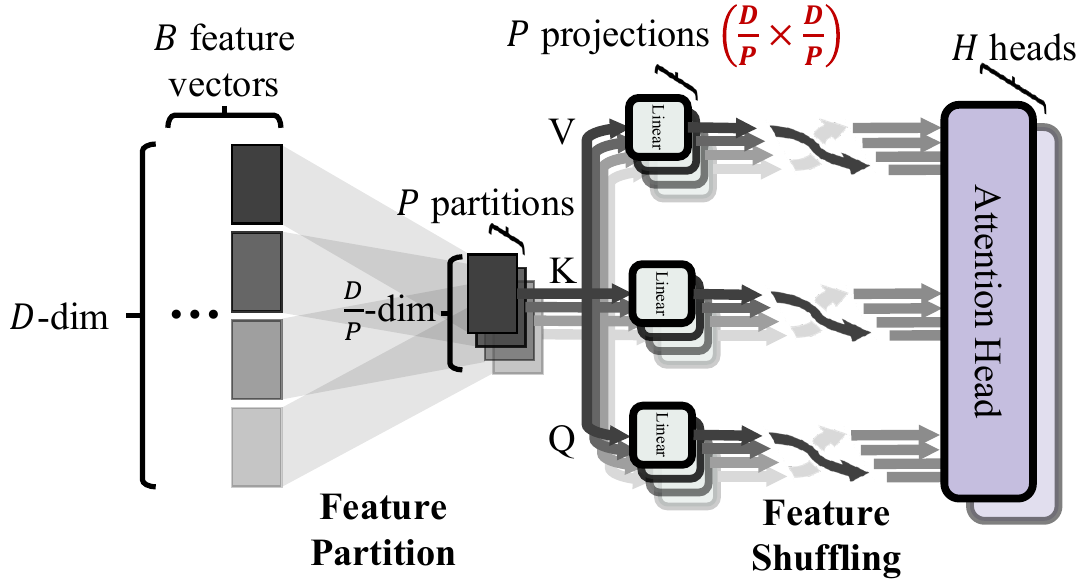}
    \caption{Partitioned linear projection with feature shuffling}
    \label{fig:projection:ps}
  \end{subfigure}
  \begin{subfigure}{0.58\columnwidth}
    \centering
    \includegraphics[width=\textwidth]{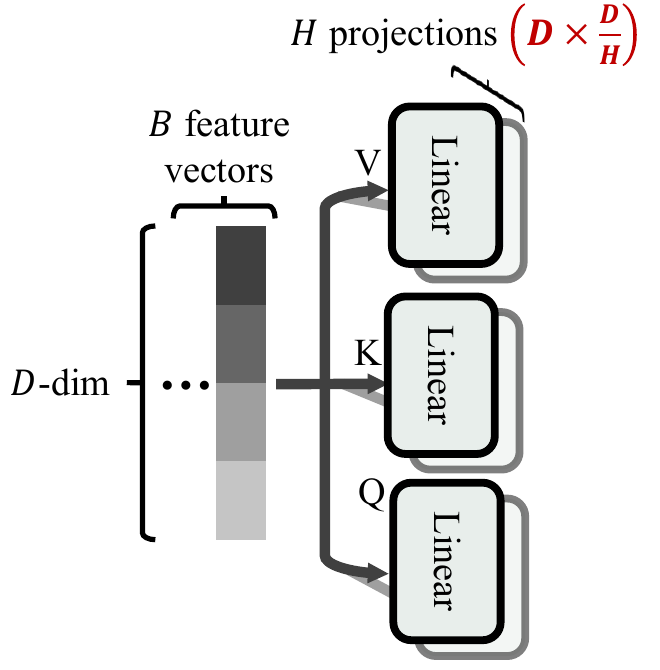}
    \caption{Without feature partition}
    \label{fig:projection:wp}
  \end{subfigure}
  \hfill
  \begin{subfigure}{0.38\columnwidth}
    \centering
    \includegraphics[width=\textwidth]{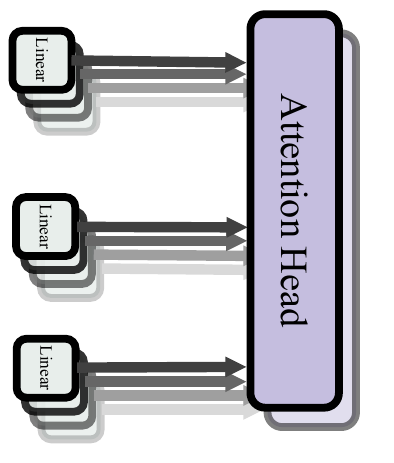}
    \caption{Without feature shuffling}
    \label{fig:projection:ws}
  \end{subfigure}
  
  \caption{
    Illustration of the partitioned linear projection and feature shuffling, and comparison with the design without feature partition or without feature shuffling.
  }\label{fig:projection}
  \vspace{-1em}
\end{figure}

\subsection{Partitioned Linear Projection}
\label{sec:approach:trick1}


The original linear projections in the \algname module requires a large size of parameters, which may become the bottleneck of cloud-based model serving for numerous clients in low latency or deploying the model to resource-constrained clients for real-time local serving. To overcome this bottleneck, we propose a partitioned version of the linear projection and allow the number of partitions to be adjusted according to the efficiency requirement. As shown in \cref{fig:projection:ps}, the input features are first partitioned into $P$ small blocks, and then a linear projection with a small size of parameters will be created for each individual block. Suppose the input feature dimension is $D$. As illustrated in \cref{fig:projection:wp}, the original linear projection requires $H\times D\times D/H = D^2$ parameters in total. In contrast, the new partitioned projection needs only $P \times (D/P)^2=D^2/P$ parameters. Therefore, the partitioned projection can reduce the number of parameters by a factor of $P$.


\subsection{Feature Shuffling}
\label{sec:approach:trick2}

Compared with the original linear projection, the new partitioned projection can improve efficiency, but isolates the features in different partitions, as depicted in \cref{fig:projection:ws}, which may hurt model performance. To mitigate side effect, we insert a feature shuffling layer after each projection layer, mixing the features from different partitions. Such a shuffling layer is implemented using a transpose operation followed by a reshape operation and introduces no additional parameters. By comparing \cref{fig:projection:ps} with \cref{fig:projection:ws}, we can observe that with feature shuffling, each attention head receives features from all the partitions, while without it, each head receives features from only a subset of partitions.

\begin{table*}[t]
  \centering
\resizebox*{1.95\columnwidth}{!}{
  \begin{tabular}{l|cc|cc|cc}
    \toprule
    & \multicolumn{2}{c|}{\bf Backbone} & \multicolumn{2}{c|}{\bf \algname-B} & \multicolumn{2}{c}{\bf \algname-T} \\
    & \#Param. & \#FLOPs & \#Param. & \#FLOPs & \#Param. & \#FLOPs  \\
    \midrule
    MobileNetV3-L~\cite{iccv/2019/howard/mobilenetv3}   & \bf 5.5M & \bf 0.23G & 30M (538\%) & 0.47G (202\%) & 0.14M {\bf(2.47\%)} & 3.8M {\bf(1.62\%)} \\
    ResNet-152~\cite{cvpr/2016/he/resnet}      & 60M & 12G & 76M (126\%) & 1.2G (10.4\%) & 0.33M {\bf(0.54\%)} & 7.9M {\bf(0.07\%)} \\
    EfficientNet-B4~\cite{icml/2019/tan/efficientnet} & 19M & 4.6G & 58M (299\%) & 0.93G (20.2\%) & 0.25M {\bf(1.32\%)} & 6.4M {\bf(0.14\%)} \\
    Swin-B~\cite{iccv/2021/liu/swin}          & 88M & 15G & 19M (21.5\%) & 0.30G (1.97\%) & 0.09M {\bf(0.10\%)} & 2.8M {\bf(0.02\%)} \\
    ConvNeXt-L~\cite{cvpr/2022/liu/convnext}      & 198M & 34G & 43M (21.5\%) & 0.68G (1.98\%) & 0.19M {\bf(0.10\%)} & 5.0M {\bf(0.01\%)} \\
    EfficientNet-B7~\cite{icml/2019/tan/efficientnet} & 66M & 39G & 118M (178\%) & 1.9G (4.86\%) & 0.50M {\bf(0.76\%)} & 11M {\bf(0.03\%)} \\
    \bottomrule
  \end{tabular}
}
  \caption{
    The size of parameters and the number of FLOPs in \algname-B and \algname-T on \imagenet with different backbone models. The percentages in the parentheses denote the relative ratios to the backbone model.
  }\label{tb:overhead}
\end{table*}

\begin{table*}[!t]
  \centering
\resizebox*{1.95\columnwidth}{!}{
  \begin{tabular}{lcccccc}
    \toprule
    \multirow{2}{*}{\bf Dataset}                & \multirow{2}{*}{\bf Task} & \multirow{2}{*}{\bf \#Clients}   & \multirow{2}{*}{\bf \#Samples}     & \multirow{2}{*}{\bf \#Classes}          & \multicolumn{2}{c}{\bf Classes per Client}                  \\ 
    \cmidrule{6-7}
                                                &                                    &                                  &                                    &                                         & mean   & stdev            \\
    \midrule
    \inaturalist~\cite{misc/2019/inaturalist19} & image classification               & 2,295                            & 193,210                            & 1,010                                   & 45.5   & 41.8              \\
    \femnist~\cite{corr/2018/caldas/leaf}       & image classification               & 3,597                            & 817,851                            & 62                                      & 55.0   & 6.6               \\
    \celeba~\cite{iccv/2015/liu/celeba}         & multi-label classification         & 9,343                            & 200,288                            & N/A                                     & N/A    & N/A              \\
    \imagenet~\cite{cvpr/2009/deng/imagenet}    & image classification               & 3,161                            & 1,331,167                          & 1,000                                   & 15.2   & 7.1               \\
    \ucf~\cite{corr/2022/soomro/ucf101}         & action recognition                 & 121                              & 13,320                             & 101                                     & 22.8   & 9.2                \\
    \bottomrule
  \end{tabular}
}
  \caption{The datasets for evaluation, the corresponding recognition tasks, as well as the global and client-level statistics of different datasets.}\label{tb:dataset}
  \vspace{-0.8em}

\end{table*}

\subsection{Overhead Analysis}
\label{sec:approach:overhead}

Suppose $N$ layers of the \algname module is adopted, and the attention scores are computed in a window of $B$ images. Each layer has $6$ partitioned linear projections, including 3 input projections for the query, key, and value vectors, 1 output projection, as well as 2 projections in the feed-forward layer, requiring $6D^2/P$ parameters and $6BD^2/P$ FLOPs. In addition, the multi-head self-attention operation requires $B^2D$ FLOPs for computing the attention scores and $B^2D$ FLOPs for computing the attention outputs. Therefore, the total size of \algname-related parameters\footnoteref{fn:overhead} is $\text{\#Param.} = \frac{6D^2}{P} \times N.$ The \algname-related computation overhead\footnote{\label{fn:overhead}The parameters and computation overhead of the residual connections and layer normalization are negligible and thus omitted for conciseness.} for recognizing one target image given $B-1$ historical images or recognizing $B$ images in a batch (e.g., for the scenario of Google Photos categorizing one client's many local images) is $\text{\#FLOPs} = \left(\frac{6BD^2}{P} + 2B^2D\right) \times N.$

To give more intuitions about the practical efficiency of \algname, we introduce two different configurations: one is the base version with the conventional linear projections (i.e., $P=1$), called \algname-B; and the other is the tiny version with $P = 256$ partitions, called \algname-T. We take 6 different backbone image recognition models to be evaluated over \imagenet, set the number of \algname layers to $N = 3$, and set the window size for computing attention scores to $B = 16$. \cref{tb:overhead} lists the detailed size of parameters and FLOPs in \algname-B and \algname-T, as well as their relative ratios to those of the backbone. We find that compared with \algname-B, \algname-T sharply reduces the parameter size and FLOPs via the partitioned linear projection, and meanwhile, introduces quite low overhead compared with the backbone, even for some light-weight networks (e.g., MobileNetV3-L~\cite{iccv/2019/howard/mobilenetv3}) that can be deployed on mobile devices.



\section{Evaluation}
\label{sec:evaluation}

\subsection{Setup}
\label{sec:evaluation:setup}

\paragraph{Datasets and recognition tasks}

We extensively evaluate \algname with 3 different tasks on 5 representative datasets, involving 9 benchmark models.



\inaturalist~\cite{misc/2019/inaturalist19} contains images of 1,010 species of plants and animals collected and verified by the users from iNaturalist\footnote{\url{www.inaturalist.org}}, a citizen science website for naturalists. The task is to recognize the species of the images. We adopt the natural user partition of FedScale~\cite{icml/2022/lai/fedscale}, allocating each image to the corresponding user who holds the rights, while splitting the user pool into 1,901 users for training and 394 ones for testing. We leave out 20\% of the users originally for training now for validation use. We take EfficientNet-B0~\cite{icml/2019/tan/efficientnet} as the recognition model.


\femnist~\cite{corr/2018/caldas/leaf} is a dataset for recognizing hand-written digits and characters, and was built by LEAF~\cite{corr/2018/caldas/leaf} through naturally partitioning Extended MNIST~\cite{ieee/1998/lecun/mnist,corr/2017/cohen/emnist} based on the writer. LEAF originally takes 90\% and 10\% of each writer's samples for on-client training and testing, respectively. We separate out 20\% of the training samples for validation. We adopt the CNN architecture officially provided by LEAF.


\celeba~\cite{iccv/2015/liu/celeba} contains face images of 10,177 celebrities, each with 40 binary attribute annotations. The task is to recognize the attributes. Since the attributes of being ``male'' or not and being ``young'' or not are normally unique and easy to be inferred for a certain user's images, we remove these 2 attributes and keep the other 38 attributes for recognition. We take LEAF's~\cite{corr/2018/caldas/leaf} natural user partition based on face ID, dividing the user pool into 8,408 users for training and 935 ones for testing. We still leave out 20\% of the users originally for training now for validation. For multi-label classification, we replace the final layer of EfficientNet-B0~\cite{icml/2019/tan/efficientnet} with multiple linear classifiers, one for each attribute.



\imagenet~\cite{cvpr/2009/deng/imagenet} contains 1,331,167 images of 1,000 classes organized under the hierarchy of WordNet~\cite{acm/1995/miller/wordnet} with 85 parent categories. We take the official train-test split and perform manual dataset splitting for different users, where each user holds roughly 324, 81, and 16 samples from a specific parent category for training, validation, and testing, respectively. We adopt MobileNetV3-L~\cite{iccv/2019/howard/mobilenetv3}, ResNet-152~\cite{cvpr/2016/he/resnet}, EfficientNet-B4~\cite{icml/2019/tan/efficientnet}, Swin-B~\cite{iccv/2021/liu/swin}, ConvNeXt-L~\cite{cvpr/2022/liu/convnext}, and EfficientNet-B7~\cite{icml/2019/tan/efficientnet}.

\ucf~\cite{corr/2022/soomro/ucf101} contains 13,320 videos of 101 action classes and 5 general/parent action categories. The manual dataset splitting is similar to that for \imagenet, where the difference is that each user takes roughly 62, 16, and 32 videos from a certain parent category for training, validation, and testing, respectively. Regarding the action recognition model, we take C3D~\cite{iccv/2015/tran/c3d}. 

We also list some statistics of the datasets in~\cref{tb:dataset}. We can observe that no matter by natural partition or manual partition, each client's local dataset involves a (normally small) subset of all the classes, validating the starting point of the model adaptiveness requirement.


\vspace{-1em}
\paragraph{Baselines and implementation details} We introduce 3 baselines, including the original backbone model, fine-tuning, and prompt tuning, for comparison.


\emph{Original backbone model} takes the mainstream recognition network architectures without the \algname module. In addition, the backbone model is optimized over the global training set (i.e., a mixture of all the clients' training sets). We initialize EfficientNets, the other models over \imagenet, and C3D, by loading the weights pre-trained over \imagenet from efficientnet-pytorch~\cite{misc/2021/efficientnet}, torchvision~\cite{nips/2019/paszke/pytorch}, and pytorch-video-recognition~\cite{misc/2019/c3d}, respectively. For \inaturalist, \celeba, and \ucf, we further tune the pre-trained models over the global training set. For \femnist, we train the CNN model from scratch.

\emph{Fine-tuning} is to let each client fine-tune the backbone model over its local training dataset and is a classical method for model adaptation. In the experiment, we fine-tune only the last layer of the backbone model (i.e., the classifier(s)) and freeze the other layers, which can function as a static feature extractor. We note that fine-tuning is not applicable to \inaturalist or \celeba, because the train-test splits of these two datasets are by users, and the users in the training dataset for fine-tuning do not appear in the testing dataset.

\emph{Prompt tuning}, as introduced in \cref{sec:related}, was an enhanced version of fine-tuning and was originally for NLP. We adapt this method to our context by associating each client with a prompt token and feed it as an extra input to the recognition models. We try and optimally set the dimension of the prompt tokens to half of the feature dimension $D/2$. 


For the settings of our \algname module, we evaluate the two configurations of \algname-B and \algname-T introduced in \cref{sec:approach:overhead}.
We try and take the optimal number of layers for each task (i.e., $N = 1$ for \femnist and \celeba, $N = 2$ for \inaturalist and \ucf, and $N = 3$ for \imagenet); and take $H = 4$ attention heads for all the tasks.




We implement all the models and learning algorithms with PyTorch (torch 1.12.1, torchvision 0.13.1). We use stochastic gradient descent (SGD) as the optimizer. By default, the learning rate is set to 0.01, and the batch size is set to 16. More detailed settings are deferred to \cref{sec:exp_sup:set}. 

\subsection{Adaptiveness to Heterogeneous Clients}

\begin{figure}[h]
  \centering
  \includegraphics[width=0.95\columnwidth]{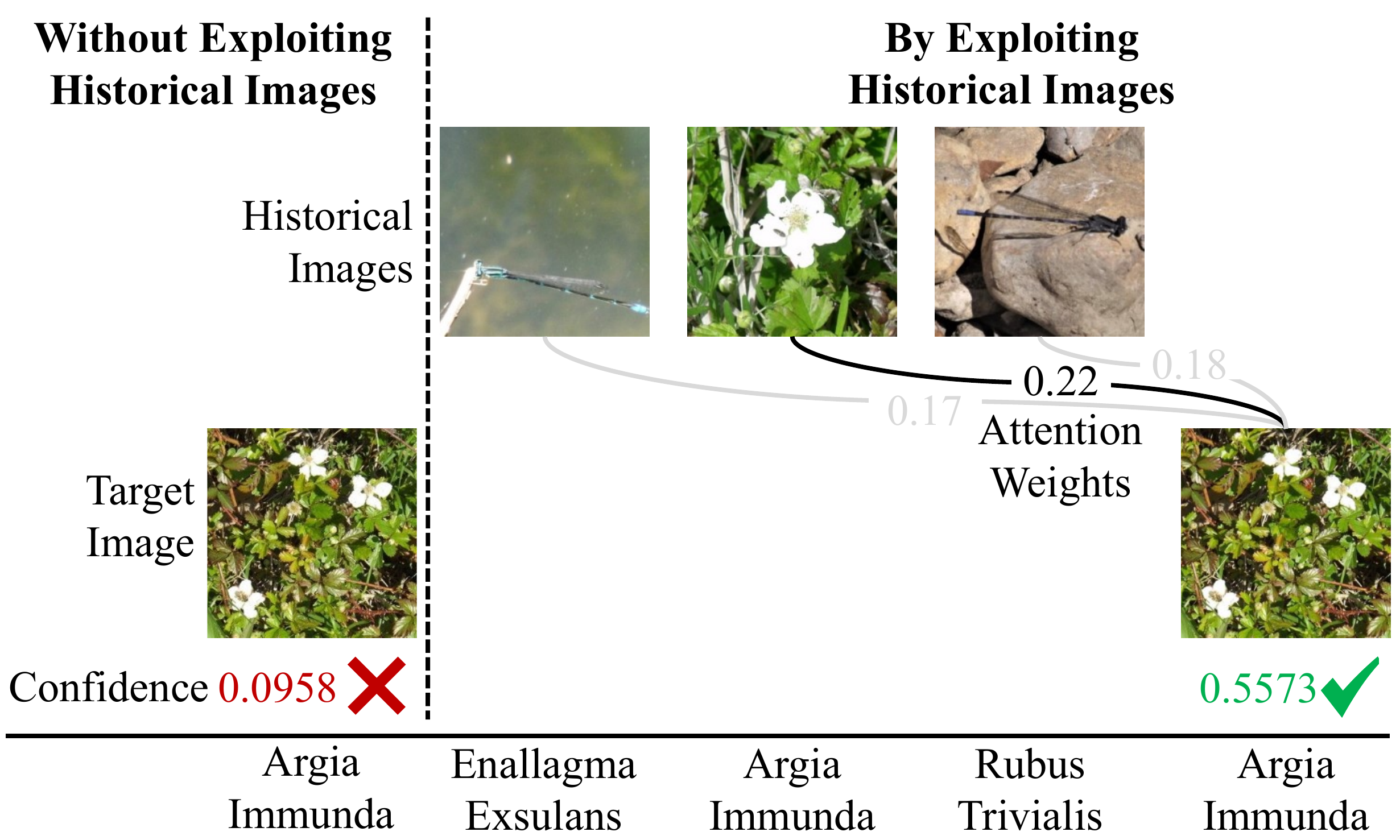}
  \caption{
    Adaptation to client \#1919 in \inaturalist.
  }\label{fig:visual}
  \vspace{-2em}
\end{figure}

\paragraph{Calibration from historical recognized images}

We first visualize why \algname can adapt to each client's local data distribution in the recognition phase. We pick client \#1919 from \inaturalist's testing set and illustrate how \algname-B calibrates the classification from historical images in \cref{fig:visual}. We can see that without exploiting any historical image, the original backbone model will misclassify the target image. In contrast, with 3 historical images, \algname-B pays the most attention to the second image from the same class (i.e., argia immunda) with the target one, improves the confidence on the target class, and classifies correctly.


\begin{figure}[h]
\centering
  \includegraphics[width=0.95\columnwidth]{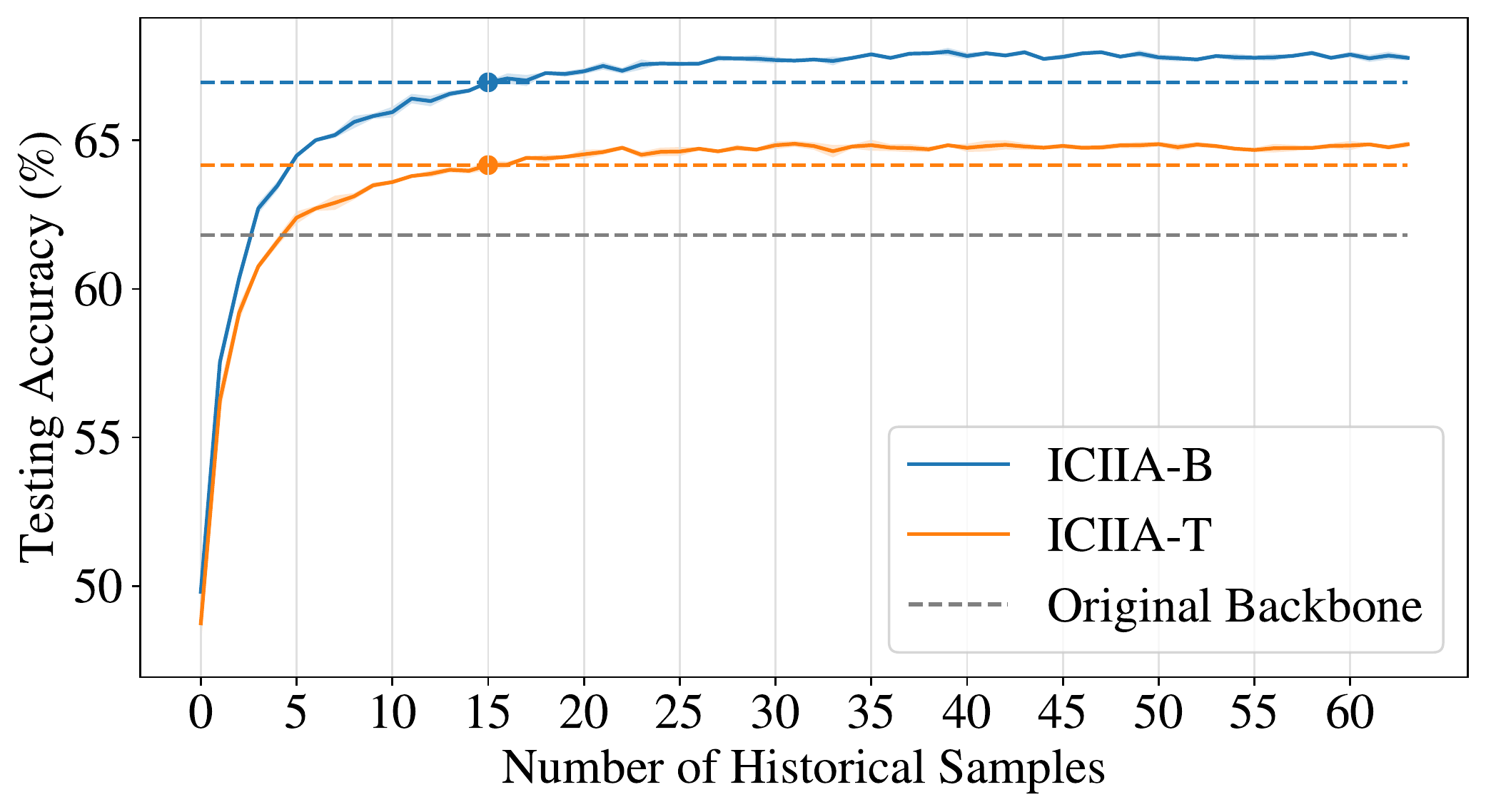}
  \vspace{-0.5em}
  \caption{
    Testing accuracy of \algname over \inaturalist by varying the number of historical samples. The dashed lines correspond to the training setting of 15 historical samples.
  }\label{fig:adapt}
  \vspace{-0.5em}
\end{figure}

We also plot how the testing accuracy of \algname-B and \algname-T changes with the accumulation of historical images in \cref{fig:adapt}. One key observation is that \algname-B and \algname-T can quickly adapt to each client's local distribution and outperform the original backbone model using only $3$ and $5$ historical images, respectively. If a client has not accumulated enough historical images yet, it can trivially switch to the original backbone model, avoiding cold start. The second key observation is that as the size of historical images increases, \algname-B and \algname-T achieve better performance. Further considering the attention is performed on the batches of 16 images (15 historical images + 1 target image) in the training phase, while \algname-B and \algname-T continues to improve after accumulating 15 historical images in the prediction phase, we can derive that \algname has a very strong generalization ability.    


\begin{figure}[h]
  \centering
  \includegraphics[width=0.95\columnwidth]{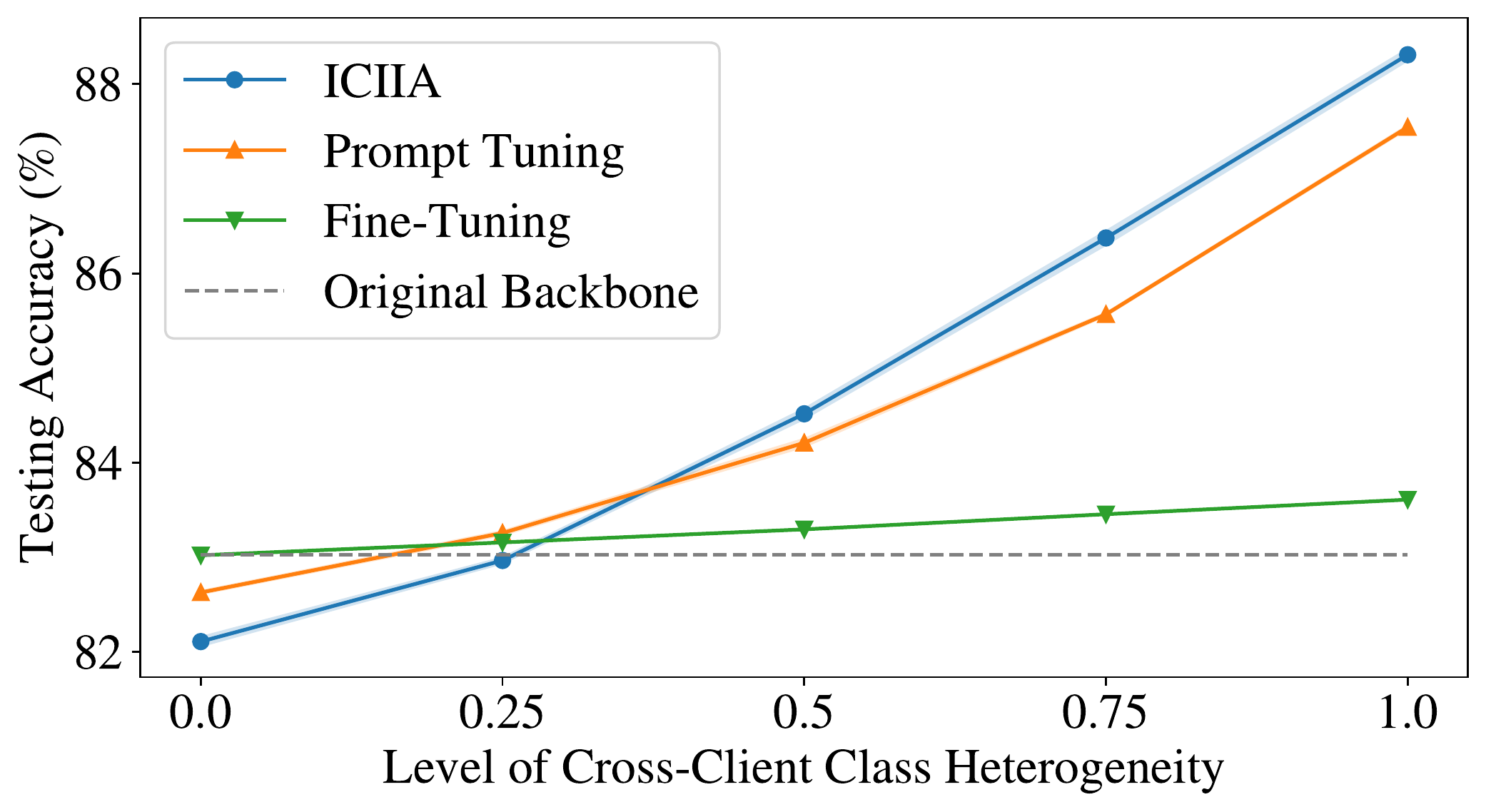}
  \vspace{-0.5em}
  \caption{
    Testing accuracy of \algname-B and three baselines over \imagenet with EfficientNet-B4, by varying the level of cross-client class heterogeneity, where the level ``0'' denotes the ideal case of homogeneity.
  }\label{fig:non-iid}
  \vspace{-1.5em}
\end{figure}

\paragraph{Adaptation to cross-client class heterogeneity} We next study how the heterogeneity of the image classes on different clients will affect \algname. We use \imagenet and vary the client-level dataset splitting. In particular, the first group of clients will now fetch the samples randomly from all the 85 parent categories, while the second group will still fetch from a specific parent category. We view the ratio between the size of the second group and the size of all the clients as a metric of cross-client class heterogeneity. 
If the ratio is 0, all the clients will fetch samples from all the parent categories, and cross-client class heterogeneity, as well as the cloud-client discrepancy, do not exist.
We depict the testing accuracy of \algname-B and the three baselines in \cref{fig:non-iid}. We can see that the advantage of \algname-B, fine-tuning, and prompt tuning over the original backbone model becomes larger as the level of cross-client class heterogeneity increases, and \algname-B performs the best. 
In the ideal case of no cross-client class heterogeneity, due to the disappearance of the device-cloud discrepancy, all the adaptive methods have no positive effect.

\begin{table*}[!t]
  \centering
   \resizebox*{1.98\columnwidth}{!}{
  \begin{tabular}{llccccc}
    \toprule
    \bf Dataset                                                 & \bf Backbone                                      & \bf Original                      & \bf Fine-Tuning                   & \bf Prompt Tuning                & \bf \algname-B             & \bf \algname-T     \\
    \midrule
    \inaturalist~\cite{misc/2019/inaturalist19}                 & EfficientNet-B0~\cite{icml/2019/tan/efficientnet} & 61.82\%                         & N/A                               & N/A                              & 66.70\% \textbf{(+4.88\%)} & 64.03\% \textbf{(+2.21\%)}               \\
    \midrule
    \femnist~\cite{corr/2018/caldas/leaf}                       & CNN~\cite{corr/2018/caldas/leaf}                  & 88.48\%                         & 90.03\%                           & 88.47\%                          & 91.94\% \textbf{(+3.46\%)} & 91.38\% \textbf{(+2.89\%)}              \\
    \midrule
    \celeba~\cite{iccv/2015/liu/celeba}                         & EfficientNet-B0~\cite{icml/2019/tan/efficientnet} & 90.84\%                         & N/A                               & N/A                              & 91.70\% \textbf{(+0.85\%)} & 91.58\% \textbf{(+0.74\%)}              \\
    \midrule
    \multirow{6}{*}{\imagenet~\cite{cvpr/2009/deng/imagenet}}   & MobileNetV3-L~\cite{iccv/2019/howard/mobilenetv3} & 75.26\%                         & 79.06\%                           & 83.93\%                          & 84.01\% \textbf{(+8.75\%)} & 83.37\% \textbf{(+8.11\%)}              \\
                                                                & ResNet-152~\cite{cvpr/2016/he/resnet}             & 82.31\%                         & 83.54\%                           & 86.85\%                          & 88.16\% \textbf{(+5.84\%)} & 87.90\% \textbf{(+5.58\%)}              \\
                                                                & EfficientNet-B4~\cite{icml/2019/tan/efficientnet} & 83.03\%                         & 83.65\%                           & 87.54\%                          & 88.31\% \textbf{(+5.28\%)} & 88.35\% \textbf{(+5.32\%)}              \\
                                                                & Swin-B~\cite{iccv/2021/liu/swin}                  & 83.58\%                         & 84.45\%                           & 87.00\%                          & 89.23\% \textbf{(+5.65\%)} & 88.86\% \textbf{(+5.28\%)}              \\
                                                                & ConvNeXt-L~\cite{cvpr/2022/liu/convnext}          & 84.42\%                         & 84.84\%                           & 87.59\%                          & 89.31\% \textbf{(+4.89\%)} & 89.14\% \textbf{(+4.72\%)}              \\
                                                                & EfficientNet-B7~\cite{icml/2019/tan/efficientnet} & 84.58\%                         & 84.98\%                           & 88.71\%                          & 89.36\% \textbf{(+4.78\%)} & 89.37\% \textbf{(+4.79\%)}              \\
    \midrule
    \ucf~\cite{corr/2022/soomro/ucf101}                         & C3D~\cite{iccv/2015/tran/c3d}                     & 79.94\%                         & 79.96\%                           & 80.01\%                          & 81.09\% \textbf{(+1.15\%)} & 80.87\% \textbf{(+0.93\%)}              \\
    \bottomrule
  \end{tabular}
   }
  \caption{
    \algname vs. the baselines in terms of testing accuracy. The improvement over the original backbone model is shown in parentheses. The results are averaged over three repeats, each repeat randomly initializing the model parameters excluding the pre-trained weights. The standard errors of the mean are all below 0.24\%.
  }
  \label{tb:comp_baseline}
\end{table*}

\begin{figure}[t]
  \centering
  \begin{subfigure}{0.95\columnwidth}
    \centering
    \includegraphics[width=\textwidth]{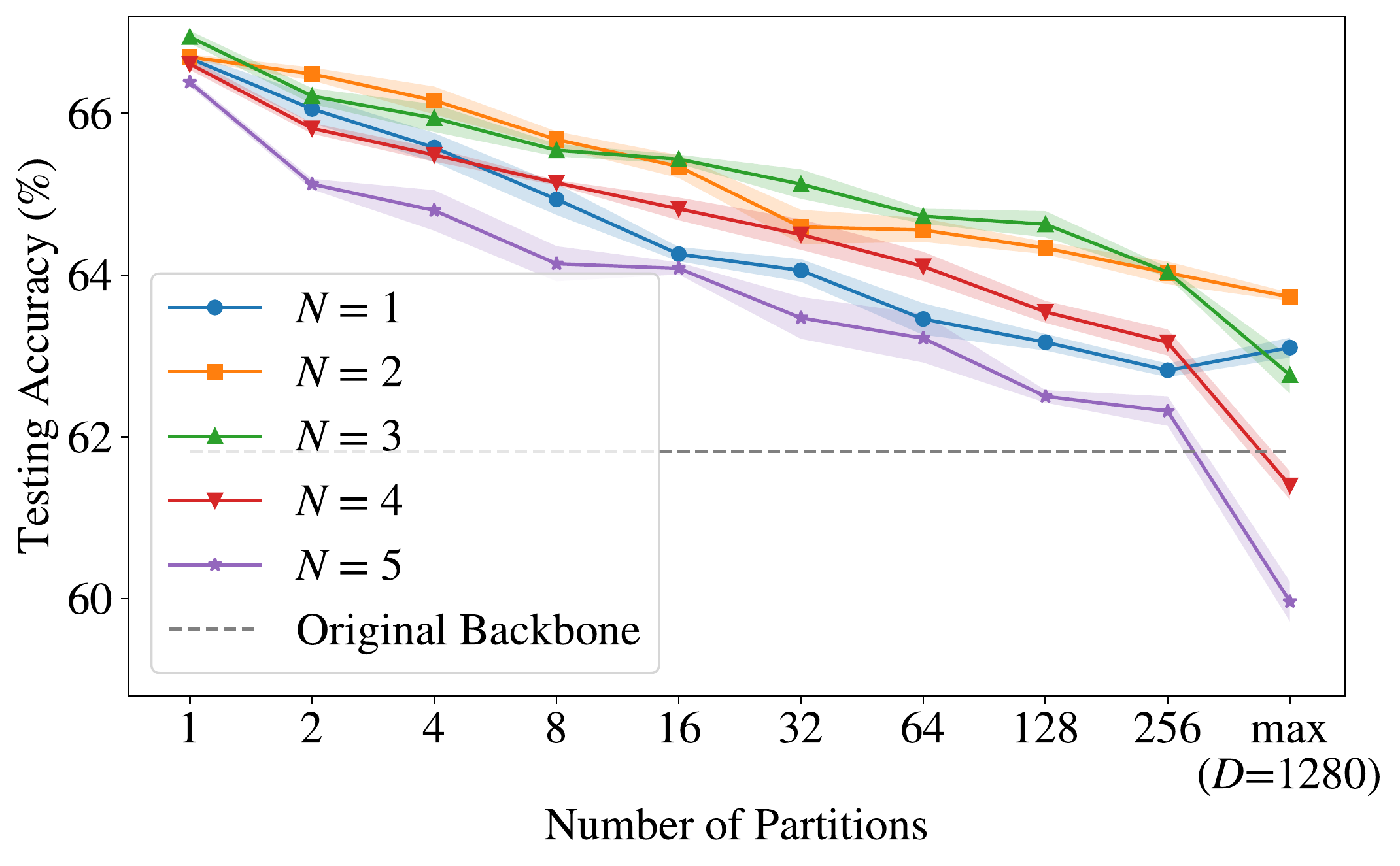}
    \caption{\inaturalist with Efficient-B0}
    \label{fig:hyper:inaturalist}
  \end{subfigure}
  \begin{subfigure}{0.95\columnwidth}
    \centering
    \includegraphics[width=\textwidth]{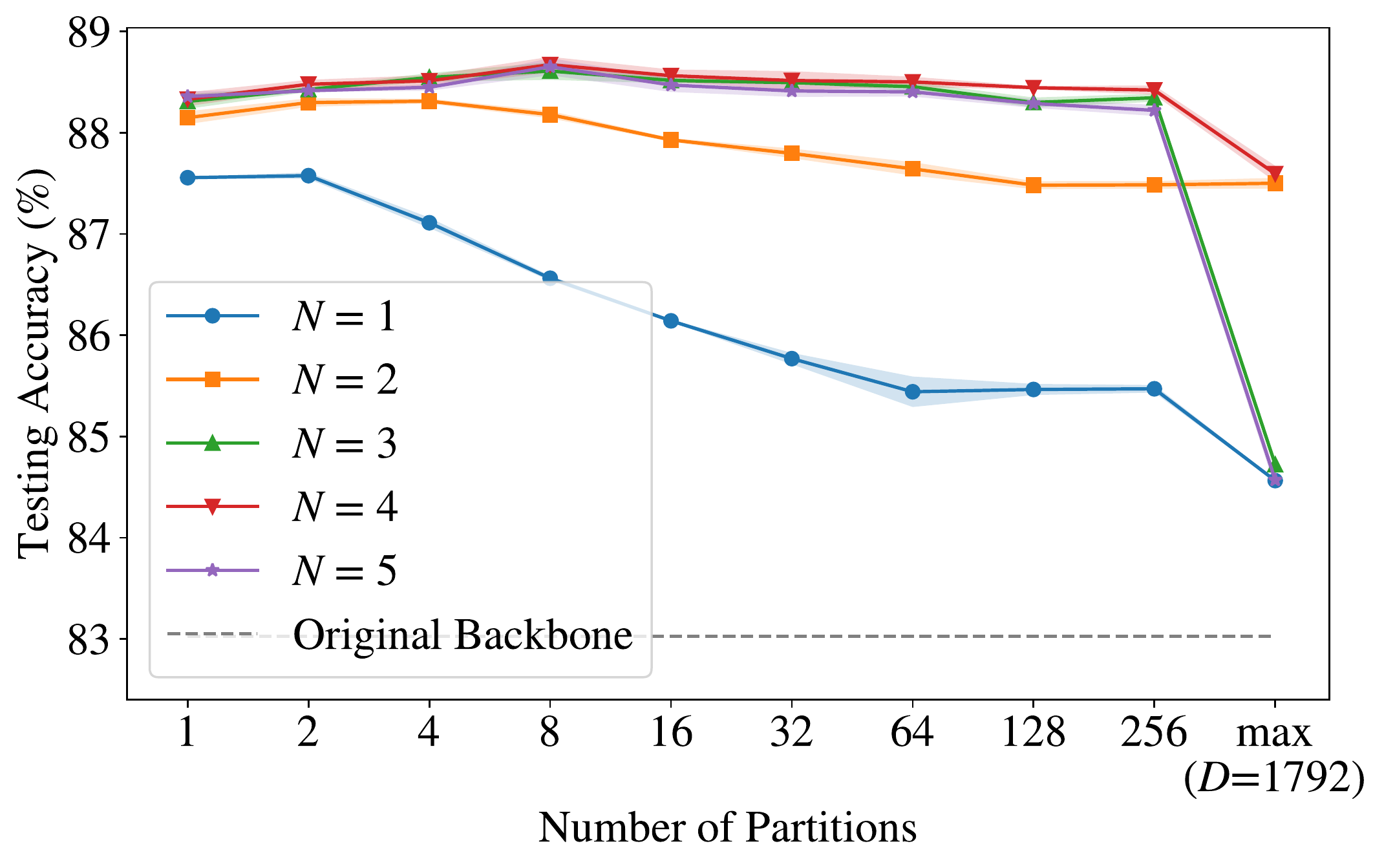}
    \caption{\imagenet with EfficientNet-B4}
    \label{fig:hyper:imagenet}
  \end{subfigure}
  \vspace{-0.4em}
  \caption{
    Testing accuracy of \algname with varying number of partitions $P$ and layers $N$. The ``max'' on x-axis denotes that $P$ equals the dimension of the input feature $D$.
    Results are averaged over three repeats, and the shaded region depicts the standard error.
  }\label{fig:hyper}
  \vspace{-1em}
\end{figure}

\vspace{-1em}
\paragraph{Adaptation of \#partitions $P$}

We finally evaluate how \algname balances model performance and efficiency by taking different numbers of partitions $P$ in partitioned linear projection. \cref{fig:hyper} shows the evaluation results over \inaturalist and \imagenet, where the number of \algname layers $N$ varies to demonstrate robustness. From \cref{fig:hyper}, we can see that as $P$ increases, the parameter size is reduced to $1/P$ of \algname-B's size, and the testing accuracy of \algname will decrease. However, even when $P$ reaches 256 (i.e., \algname-T), the testing accuracy of \algname-T is still higher than that of the original backbone model. The advantage and the robustness of \algname are more evident over \imagenet.  


\subsection{Comparison with Baselines}
\label{sec:evaluation:baselines}


We compare \algname with the baselines and report the testing accuracy in \cref{tb:comp_baseline}. We can observe consistent and significant improvements of \algname over all the baselines on all the datasets and recognition models. (1) By comparing \algname-B with the original backbone model, we can draw that \algname indeed improves model performance by exploiting each client's class heterogeneity from the local samples to be recognized; (2) by comparing \algname-B with fine-tuning and prompt tuning, we can derive that the intra-client inter-image attention mechanism has the strongest adaptiveness, even without on-client retraining; and (3) by comparing \algname-T with \algname-B and the baselines, we can draw that although \algname-T compromises model accuracy for efficiency, it still significantly outperforms the baselines.


\subsection{Ablation Study}
\label{sec:evaluation:ablation}
\begin{table}[h]
  \centering
  \resizebox*{\columnwidth}{!}{
  \begin{tabular}{lccc}
    \toprule
    \bf \multirow{2}{*}{Dataset} & \bf \algname-B    & \bf \algname-T     & \bf \algname-T \\
                                 & \bf -attention & \bf -projection & \bf -shuffling \\
    \midrule
    \inaturalist~\cite{misc/2019/inaturalist19} & -4.78\%           & -0.30\%            & -0.20\% \\
    \imagenet~\cite{cvpr/2009/deng/imagenet}    & -6.50\%           & -3.62\%            & -0.28\% \\
    \bottomrule
  \end{tabular}
  }
  \caption{
    Drop of testing accuracy after removing different ingredients of \algname. 
  }
  \vspace{-1em}
  \label{tb:ablation}
\end{table}

\paragraph{Impact of intra-client and inter-image attention}
We remove the intra-client and inter-image attention operation of \algname-B and reserve all the linear projections in both the original multi-head self-attention and feed-forward layers. As shown in the first column of \cref{tb:ablation}, the testing accuracy sharply drops by 4.78\% on \inaturalist and by 6.50\% on \imagenet, which validates the necessity of intra-client and inter-image attention.


\vspace{-1em}
\paragraph{Impact of partitioned linear projection} We remove the partitioned linear projections of \algname-T and reserve only the intra-client and inter-image attention operation. From the second column of \cref{tb:ablation}, we can see that the testing accuracy slightly drops on \inaturalist, but dramatically drops by 3.62\% on \imagenet, validating the necessity of partitioned linear projections.

%

\begin{figure}[t]
  \centering
  \begin{subfigure}{0.9\columnwidth}
    \centering
    \includegraphics[width=\textwidth]{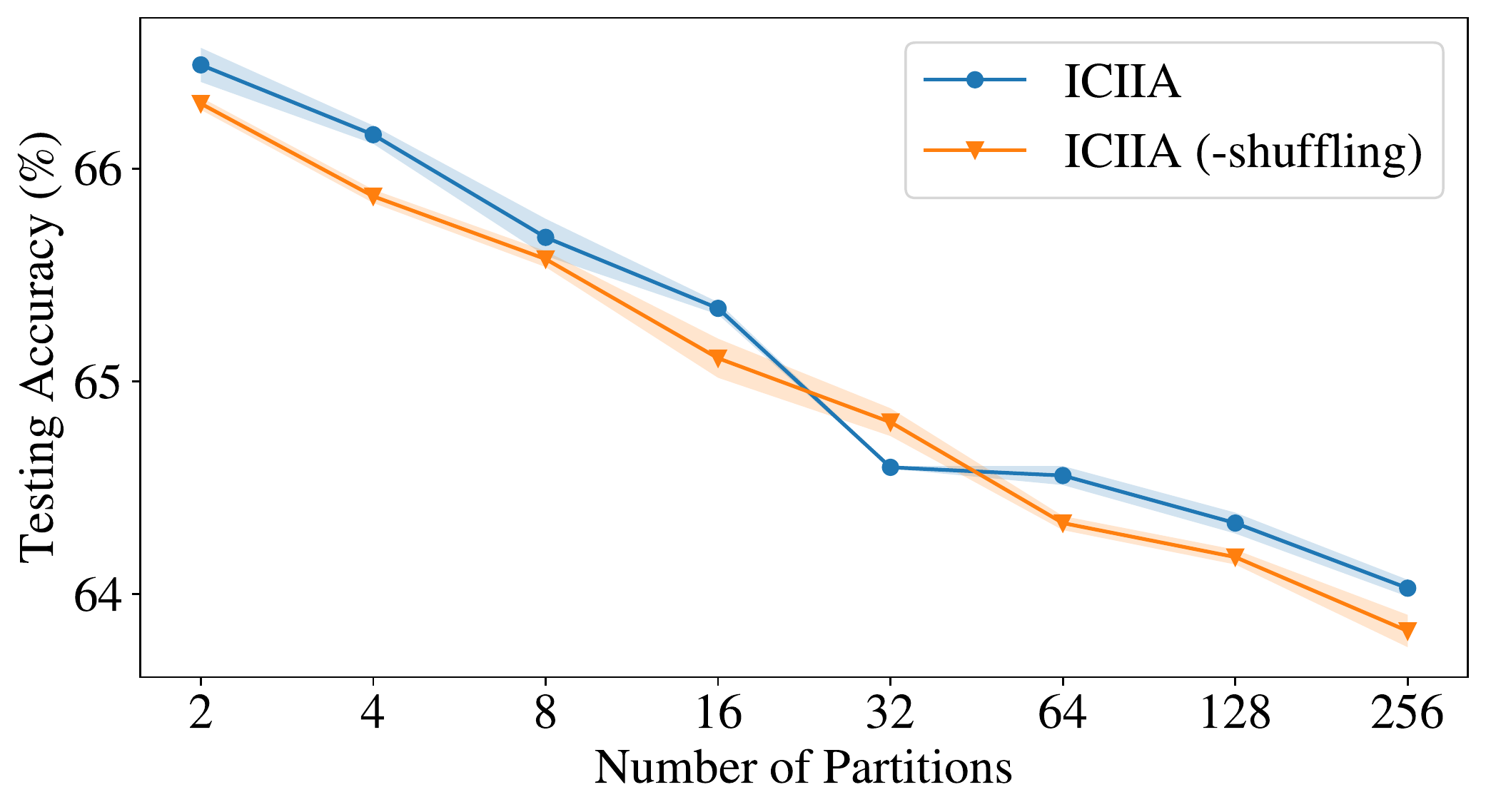}
    \caption{\inaturalist with EfficientNet-B0}
    \label{fig:shuffle:inaturalist}
  \end{subfigure}
  \begin{subfigure}{0.9\columnwidth}
    \centering
    \includegraphics[width=\textwidth]{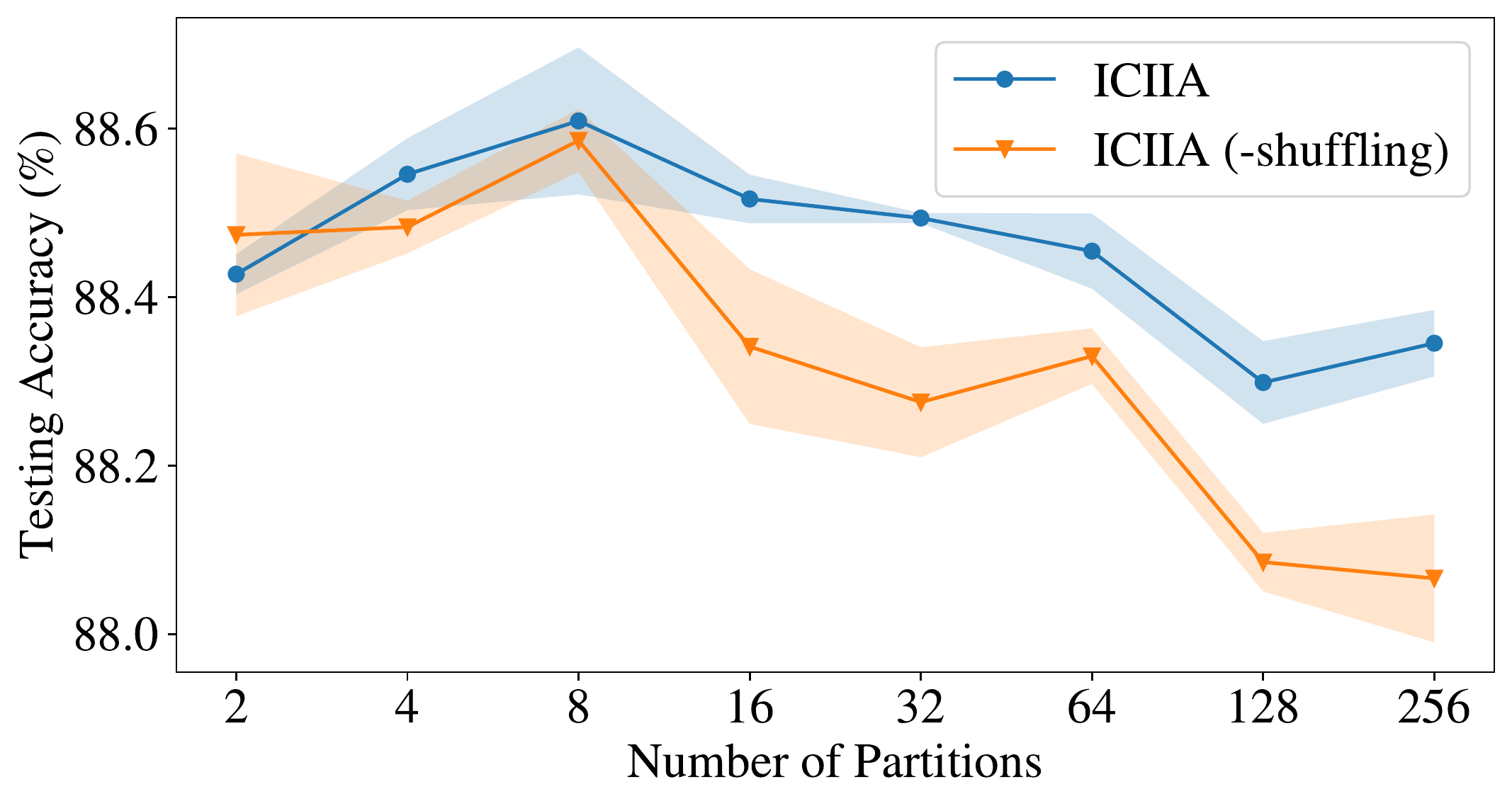}
    \caption{\imagenet with EfficientNet-B4}
    \label{fig:shuffle:imagenet}
  \end{subfigure}
  \vspace{-0.5em}
  \caption{\algname with and without feature shuffling, by varying the number of partitions $P$.}
  \label{fig:shuffle}
  \vspace{-1em}
\end{figure}

\vspace{-1em}
\paragraph{Impact of feature shuffling} We remove the feature shuffling operation from \algname-T to ablate its impact. We can see from the third column of \cref{tb:ablation} that the testing accuracy slightly drops. We also evaluate \algname with and without feature shuffling by varying the number of partitions $P$ in \cref{fig:shuffle}. We can see that the drop of accuracy generally becomes larger for a larger $P$. These results validate the necessity of feature shuffling, especially together with a large number of partitioned projections.



\section{Conclusion}
\label{sec:conclusion}
\vspace{-0.2em}
In this work, we study the ubiquitous image recognition applications from a new cloud-client perspective. We have proposed a plugable \algname module to adapt the backbone recognition model from the cloud's full set of classes to each individual client's local dynamic subset of classes, simply after one-time cloud-based training. \algname captures intra-client and inter-image dependencies with multi-head self-attention and further improves efficiency by partitioned linear projection along with feature shuffling. We have extensively evaluated \algname, revealing effectiveness, efficiency, and superiority. 

\clearpage
{\small
\bibliographystyle{ieee_fullname}
\bibliography{main}

\begin{thebibliography}{10}\itemsep=-1pt

\bibitem{corr/2016/ba/layernorm}
Lei~Jimmy Ba, Jamie~Ryan Kiros, and Geoffrey~E. Hinton.
\newblock Layer normalization.
\newblock {\em CoRR}, abs/1607.06450, 2016.

\bibitem{nips/2020/brown/gpt3}
Tom~B. Brown, Benjamin Mann, Nick Ryder, Melanie Subbiah, Jared Kaplan,
  Prafulla Dhariwal, Arvind Neelakantan, Pranav Shyam, Girish Sastry, Amanda
  Askell, Sandhini Agarwal, Ariel Herbert{-}Voss, Gretchen Krueger, Tom
  Henighan, Rewon Child, Aditya Ramesh, Daniel~M. Ziegler, Jeffrey Wu, Clemens
  Winter, Christopher Hesse, Mark Chen, Eric Sigler, Mateusz Litwin, Scott
  Gray, Benjamin Chess, Jack Clark, Christopher Berner, Sam McCandlish, Alec
  Radford, Ilya Sutskever, and Dario Amodei.
\newblock Language models are few-shot learners.
\newblock In {\em {NeurIPS}}, 2020.

\bibitem{nips/2020/cai/tinytl}
Han Cai, Chuang Gan, Ligeng Zhu, and Song Han.
\newblock Tinytl: Reduce memory, not parameters for efficient on-device
  learning.
\newblock In {\em {NeurIPS}}, 2020.

\bibitem{corr/2018/caldas/leaf}
Sebastian Caldas, Peter Wu, Tian Li, Jakub Kone{\v{c}}n{\'y}, H.~Brendan
  McMahan, Virginia Smith, and Ameet Talwalkar.
\newblock {LEAF:} {A} benchmark for federated settings.
\newblock {\em CoRR}, abs/1812.01097, 2018.

\bibitem{corr/2018/chen/meta}
Fei Chen, Mi Luo, Zhenhua Dong, Zhenguo Li, and Xiuqiang He.
\newblock Federated meta-learning with fast convergence and efficient
  communication.
\newblock {\em CoRR}, abs/1802.07876, 2018.

\bibitem{cvpr/2022/chen/visualgpt}
Jun Chen, Han Guo, Kai Yi, Boyang Li, and Mohamed Elhoseiny.
\newblock Visualgpt: Data-efficient adaptation of pretrained language models
  for image captioning.
\newblock In {\em {CVPR}}, 2022.

\bibitem{corr/2017/cohen/emnist}
Gregory Cohen, Saeed Afshar, Jonathan Tapson, and Andr{\'{e}} van Schaik.
\newblock {EMNIST:} an extension of {MNIST} to handwritten letters.
\newblock {\em CoRR}, abs/1702.05373, 2017.

\bibitem{cvpr/2009/deng/imagenet}
Jia Deng, Wei Dong, Richard Socher, Li{-}Jia Li, Kai Li, and Li Fei{-}Fei.
\newblock Imagenet: {A} large-scale hierarchical image database.
\newblock In {\em {CVPR}}, 2009.

\bibitem{naacl/devlin/2019/bert}
Jacob Devlin, Ming{-}Wei Chang, Kenton Lee, and Kristina Toutanova.
\newblock {BERT:} pre-training of deep bidirectional transformers for language
  understanding.
\newblock In {\em {NAACL-HLT} {(1)}}, 2019.

\bibitem{corr/2022/ding/label}
Yucheng Ding, Chaoyue Niu, Fan Wu, Shaojie Tang, Chengfei Lyu, and Guihai Chen.
\newblock On-device model fine-tuning with label correction in recommender
  systems.
\newblock {\em CoRR}, abs/2211.01163, 2022.

\bibitem{nips/2020/dinh/envelope}
Canh~T. Dinh, Nguyen~Hoang Tran, and Tuan~Dung Nguyen.
\newblock Personalized federated learning with moreau envelopes.
\newblock In {\em {NeurIPS}}, 2020.

\bibitem{iclr/2021/dosovitskiy/vit}
Alexey Dosovitskiy, Lucas Beyer, Alexander Kolesnikov, Dirk Weissenborn,
  Xiaohua Zhai, Thomas Unterthiner, Mostafa Dehghani, Matthias Minderer, Georg
  Heigold, Sylvain Gelly, Jakob Uszkoreit, and Neil Houlsby.
\newblock An image is worth 16x16 words: Transformers for image recognition at
  scale.
\newblock In {\em {ICLR}}, 2021.

\bibitem{cvpr/2022/du/l2p}
Yu Du, Fangyun Wei, Zihe Zhang, Miaojing Shi, Yue Gao, and Guoqi Li.
\newblock Learning to prompt for open-vocabulary object detection with
  vision-language model.
\newblock In {\em {CVPR}}, 2022.

\bibitem{cvpr/2022/du/cocoop}
Yu Du, Fangyun Wei, Zihe Zhang, Miaojing Shi, Yue Gao, and Guoqi Li.
\newblock Learning to prompt for open-vocabulary object detection with
  vision-language model.
\newblock In {\em {CVPR}}, 2022.

\bibitem{nips/2020/fallah/flmaml}
Alireza Fallah, Aryan Mokhtari, and Asuman~E. Ozdaglar.
\newblock Personalized federated learning with theoretical guarantees: {A}
  model-agnostic meta-learning approach.
\newblock In {\em {NeurIPS}}, 2020.

\bibitem{cvpr/2019/guo/spottune}
Yunhui Guo, Honghui Shi, Abhishek Kumar, Kristen Grauman, Tajana Rosing, and
  Rog{\'{e}}rio~Schmidt Feris.
\newblock Spottune: Transfer learning through adaptive fine-tuning.
\newblock In {\em {CVPR}}, 2019.

\bibitem{cvpr/2022/he/mae}
Kaiming He, Xinlei Chen, Saining Xie, Yanghao Li, Piotr Doll{\'{a}}r, and
  Ross~B. Girshick.
\newblock Masked autoencoders are scalable vision learners.
\newblock In {\em {CVPR}}, 2022.

\bibitem{cvpr/2020/he/probe}
Kaiming He, Haoqi Fan, Yuxin Wu, Saining Xie, and Ross~B. Girshick.
\newblock Momentum contrast for unsupervised visual representation learning.
\newblock In {\em {CVPR}}, 2020.

\bibitem{cvpr/2016/he/resnet}
Kaiming He, Xiangyu Zhang, Shaoqing Ren, and Jian Sun.
\newblock Deep residual learning for image recognition.
\newblock In {\em {CVPR}}, 2016.

\bibitem{misc/2019/inaturalist19}
Grant~Van Horn and Oisin~Mac Aodha.
\newblock {{iNaturalist 2019}}.
\newblock
  \url{https://sites.google.com/view/fgvc6/competitions/inaturalist-2019},
  2019.

\bibitem{icml/2019/houlsby/adapter}
Neil Houlsby, Andrei Giurgiu, Stanislaw Jastrzebski, Bruna Morrone, Quentin de
  Laroussilhe, Andrea Gesmundo, Mona Attariyan, and Sylvain Gelly.
\newblock Parameter-efficient transfer learning for {NLP}.
\newblock In {\em {ICML}}, 2019.

\bibitem{iccv/2019/howard/mobilenetv3}
Andrew Howard, Ruoming Pang, Hartwig Adam, Quoc~V. Le, Mark Sandler, Bo Chen,
  Weijun Wang, Liang{-}Chieh Chen, Mingxing Tan, Grace Chu, Vijay Vasudevan,
  and Yukun Zhu.
\newblock Searching for mobilenetv3.
\newblock In {\em {ICCV}}, 2019.

\bibitem{iclr/2022/hu/lora}
Edward~J. Hu, Yelong Shen, Phillip Wallis, Zeyuan Allen{-}Zhu, Yuanzhi Li,
  Shean Wang, Lu Wang, and Weizhu Chen.
\newblock Lora: Low-rank adaptation of large language models.
\newblock In {\em {ICLR}}, 2022.

\bibitem{icml/2021/jia/align}
Chao Jia, Yinfei Yang, Ye Xia, Yi{-}Ting Chen, Zarana Parekh, Hieu Pham,
  Quoc~V. Le, Yun{-}Hsuan Sung, Zhen Li, and Tom Duerig.
\newblock Scaling up visual and vision-language representation learning with
  noisy text supervision.
\newblock In {\em {ICML}}, 2021.

\bibitem{eccv/2022/jia/vpt}
Menglin Jia, Luming Tang, Bor{-}Chun Chen, Claire Cardie, Serge~J. Belongie,
  Bharath Hariharan, and Ser{-}Nam Lim.
\newblock Visual prompt tuning.
\newblock In {\em {ECCV} {(33)}}, 2022.

\bibitem{corr/2019/jiang/flmaml}
Yihan Jiang, Jakub Kone{\v{c}}n{\'y}, Keith Rush, and Sreeram Kannan.
\newblock Improving federated learning personalization via model agnostic meta
  learning.
\newblock {\em CoRR}, abs/1909.12488, 2019.

\bibitem{icml/2022/lai/fedscale}
Fan Lai, Yinwei Dai, Sanjay Sri~Vallabh Singapuram, Jiachen Liu, Xiangfeng Zhu,
  Harsha~V. Madhyastha, and Mosharaf Chowdhury.
\newblock Fedscale: Benchmarking model and system performance of federated
  learning at scale.
\newblock In {\em {ICML}}, 2022.

\bibitem{ieee/1998/lecun/mnist}
Yann LeCun, L{\'{e}}on Bottou, Yoshua Bengio, and Patrick Haffner.
\newblock Gradient-based learning applied to document recognition.
\newblock {\em Proceedings of the {IEEE}}, 86(11):2278--2324, 1998.

\bibitem{emnlp/2021/lester/prompt}
Brian Lester, Rami Al{-}Rfou, and Noah Constant.
\newblock The power of scale for parameter-efficient prompt tuning.
\newblock In {\em {EMNLP} {(1)}}, 2021.

\bibitem{corr/2020/li/lotteryfl}
Ang Li, Jingwei Sun, Binghui Wang, Lin Duan, Sicheng Li, Yiran Chen, and Hai
  Li.
\newblock Lotteryfl: Personalized and communication-efficient federated
  learning with lottery ticket hypothesis on non-iid datasets.
\newblock {\em CoRR}, abs/2008.03371, 2020.

\bibitem{icml/2021/li/ditto}
Tian Li, Shengyuan Hu, Ahmad Beirami, and Virginia Smith.
\newblock Ditto: Fair and robust federated learning through personalization.
\newblock In {\em {ICML}}, 2021.

\bibitem{cvpr/2022/li/domain}
Wei{-}Hong Li, Xialei Liu, and Hakan Bilen.
\newblock Cross-domain few-shot learning with task-specific adapters.
\newblock In {\em {CVPR}}, 2022.

\bibitem{icml/2018/li/explicit}
Xuhong Li, Yves Grandvalet, and Franck Davoine.
\newblock Explicit inductive bias for transfer learning with convolutional
  networks.
\newblock In {\em {ICML}}, 2018.

\bibitem{acl/2021/li/prefix}
Xiang~Lisa Li and Percy Liang.
\newblock Prefix-tuning: Optimizing continuous prompts for generation.
\newblock In {\em {ACL/IJCNLP} {(1)}}, 2021.

\bibitem{corr/2022/lian/ssl}
Dongze Lian, Daquan Zhou, Jiashi Feng, and Xinchao Wang.
\newblock Scaling {\&} shifting your features: {A} new baseline for efficient
  model tuning.
\newblock {\em CoRR}, abs/2210.08823, 2022.

\bibitem{corr/2020/liang/lg}
Paul~Pu Liang, Terrance Liu, Ziyin Liu, Ruslan Salakhutdinov, and
  Louis{-}Philippe Morency.
\newblock Think locally, act globally: Federated learning with local and global
  representations.
\newblock {\em CoRR}, abs/2001.01523, 2020.

\bibitem{iccv/2021/liu/swin}
Ze Liu, Yutong Lin, Yue Cao, Han Hu, Yixuan Wei, Zheng Zhang, Stephen Lin, and
  Baining Guo.
\newblock Swin transformer: Hierarchical vision transformer using shifted
  windows.
\newblock In {\em {ICCV}}, 2021.

\bibitem{iccv/2015/liu/celeba}
Ziwei Liu, Ping Luo, Xiaogang Wang, and Xiaoou Tang.
\newblock Deep learning face attributes in the wild.
\newblock In {\em {ICCV}}, 2015.

\bibitem{cvpr/2022/liu/convnext}
Zhuang Liu, Hanzi Mao, Chao{-}Yuan Wu, Christoph Feichtenhofer, Trevor Darrell,
  and Saining Xie.
\newblock A convnet for the 2020s.
\newblock In {\em {CVPR}}, 2022.

\bibitem{cvpr/2022/liu/swin}
Ze Liu, Jia Ning, Yue Cao, Yixuan Wei, Zheng Zhang, Stephen Lin, and Han Hu.
\newblock Video swin transformer.
\newblock In {\em {CVPR}}, 2022.

\bibitem{cvpr/2022/lu/prompt}
Yuning Lu, Jianzhuang Liu, Yonggang Zhang, Yajing Liu, and Xinmei Tian.
\newblock Prompt distribution learning.
\newblock In {\em {CVPR}}, 2022.

\bibitem{nips/2021/luo/fear}
Mi Luo, Fei Chen, Dapeng Hu, Yifan Zhang, Jian Liang, and Jiashi Feng.
\newblock No fear of heterogeneity: Classifier calibration for federated
  learning with non-iid data.
\newblock In {\em {NeurIPS}}, 2021.

\bibitem{cvpr/2022/ma/layer}
Xiaosong Ma, Jie Zhang, Song Guo, and Wenchao Xu.
\newblock Layer-wised model aggregation for personalized federated learning.
\newblock In {\em {CVPR}}, 2022.

\bibitem{nips/2021/mahabadi/compacter}
Rabeeh~Karimi Mahabadi, James Henderson, and Sebastian Ruder.
\newblock Compacter: Efficient low-rank hypercomplex adapter layers.
\newblock In {\em {NeurIPS}}, 2021.

\bibitem{acl/2021/mahabadi/hyper}
Rabeeh~Karimi Mahabadi, Sebastian Ruder, Mostafa Dehghani, and James Henderson.
\newblock Parameter-efficient multi-task fine-tuning for transformers via
  shared hypernetworks.
\newblock In {\em {ACL/IJCNLP} {(1)}}, 2021.

\bibitem{corr/2020/mansour/three}
Yishay Mansour, Mehryar Mohri, Jae Ro, and Ananda~Theertha Suresh.
\newblock Three approaches for personalization with applications to federated
  learning.
\newblock {\em CoRR}, abs/2002.10619, 2020.

\bibitem{aistats/2017/mcmahan/fl}
Brendan McMahan, Eider Moore, Daniel Ramage, Seth Hampson, and
  Blaise~Ag{\"{u}}era y Arcas.
\newblock Communication-efficient learning of deep networks from decentralized
  data.
\newblock In {\em {AISTATS}}, 2017.

\bibitem{misc/2021/efficientnet}
Luke Melas-Kyriazi.
\newblock {EfficientNet-PyTorch}.
\newblock \url{https://github.com/lukemelas/EfficientNet-PyTorch}, 2021.

\bibitem{acm/1995/miller/wordnet}
George~A. Miller.
\newblock Wordnet: {A} lexical database for english.
\newblock {\em Communications of the {ACM}}, 38(11):39--41, 1995.

\bibitem{nips/2019/paszke/pytorch}
Adam Paszke, Sam Gross, Francisco Massa, Adam Lerer, James Bradbury, Gregory
  Chanan, Trevor Killeen, Zeming Lin, Natalia Gimelshein, Luca Antiga, Alban
  Desmaison, Andreas K{\"{o}}pf, Edward~Z. Yang, Zachary DeVito, Martin Raison,
  Alykhan Tejani, Sasank Chilamkurthy, Benoit Steiner, Lu Fang, Junjie Bai, and
  Soumith Chintala.
\newblock Pytorch: An imperative style, high-performance deep learning library.
\newblock In {\em {NeurIPS}}, 2019.

\bibitem{icml/2021/radford/clip}
Alec Radford, Jong~Wook Kim, Chris Hallacy, Aditya Ramesh, Gabriel Goh,
  Sandhini Agarwal, Girish Sastry, Amanda Askell, Pamela Mishkin, Jack Clark,
  Gretchen Krueger, and Ilya Sutskever.
\newblock Learning transferable visual models from natural language
  supervision.
\newblock In {\em {ICML}}, 2021.

\bibitem{cvpr/2022/shen/pfed}
Yiqing Shen, Yuyin Zhou, and Lequan Yu.
\newblock Cd\({}^{\mbox{2}}\)-pfed: Cyclic distillation-guided channel
  decoupling for model personalization in federated learning.
\newblock In {\em {CVPR}}, 2022.

\bibitem{nips/2017/smith/flmulti}
Virginia Smith, Chao{-}Kai Chiang, Maziar Sanjabi, and Ameet Talwalkar.
\newblock Federated multi-task learning.
\newblock In {\em {NeurIPS}}, 2017.

\bibitem{corr/2022/soomro/ucf101}
Khurram Soomro, Amir~Roshan Zamir, and Mubarak Shah.
\newblock {UCF101:} {A} dataset of 101 human actions classes from videos in the
  wild.
\newblock {\em CoRR}, abs/1212.0402, 2012.

\bibitem{cvpr/2022/sung/vladapter}
Yi{-}Lin Sung, Jaemin Cho, and Mohit Bansal.
\newblock {VL-ADAPTER:} parameter-efficient transfer learning for
  vision-and-language tasks.
\newblock In {\em {CVPR}}, 2022.

\bibitem{icml/2019/tan/efficientnet}
Mingxing Tan and Quoc~V. Le.
\newblock Efficientnet: Rethinking model scaling for convolutional neural
  networks.
\newblock In {\em {ICML}}, 2019.

\bibitem{iccv/2015/tran/c3d}
Du Tran, Lubomir~D. Bourdev, Rob Fergus, Lorenzo Torresani, and Manohar Paluri.
\newblock Learning spatiotemporal features with 3d convolutional networks.
\newblock In {\em {ICCV}}, 2015.

\bibitem{nips/2017/vaswani/transformer}
Ashish Vaswani, Noam Shazeer, Niki Parmar, Jakob Uszkoreit, Llion Jones,
  Aidan~N. Gomez, Lukasz Kaiser, and Illia Polosukhin.
\newblock Attention is all you need.
\newblock In {\em {NeurIPS}}, 2017.

\bibitem{cvpr/2022/wang/domain}
Fan Wang, Zhongyi Han, Yongshun Gong, and Yilong Yin.
\newblock Exploring domain-invariant parameters for source free domain
  adaptation.
\newblock In {\em {CVPR}}, 2022.

\bibitem{kdd/2022/yan/mpda}
Yikai Yan, Chaoyue Niu, Renjie Gu, Fan Wu, Shaojie Tang, Lifeng Hua, Chengfei
  Lyu, and Guihai Chen.
\newblock On-device learning for model personalization with large-scale
  cloud-coordinated domain adaption.
\newblock In {\em {KDD}}, 2022.

\bibitem{cvpr/2022/ye/night}
Junjie Ye, Changhong Fu, Guangze Zheng, Danda~Pani Paudel, and Guang Chen.
\newblock Unsupervised domain adaptation for nighttime aerial tracking.
\newblock In {\em {CVPR}}, 2022.

\bibitem{aaai/2021/yu/meta}
Runsheng Yu, Yu Gong, Xu He, Yu Zhu, Qingwen Liu, Wenwu Ou, and Bo An.
\newblock Personalized adaptive meta learning for cold-start user preference
  prediction.
\newblock In {\em {AAAI}}, 2021.

\bibitem{acl/2022/zaken/bitfit}
Elad~Ben Zaken, Yoav Goldberg, and Shauli Ravfogel.
\newblock Bitfit: Simple parameter-efficient fine-tuning for transformer-based
  masked language-models.
\newblock In {\em {ACL} {(2)}}, 2022.

\bibitem{misc/2019/c3d}
Jianfeng Zhang.
\newblock {Pytorch-Video-Recognition}.
\newblock \url{https://github.com/jfzhang95/pytorch-video-recognition}, 2019.

\bibitem{cvpr/2022/zhang/flkd}
Lin Zhang, Li Shen, Liang Ding, Dacheng Tao, and Ling{-}Yu Duan.
\newblock Fine-tuning global model via data-free knowledge distillation for
  non-iid federated learning.
\newblock In {\em {CVPR}}, 2022.

\bibitem{aaai/2019/zhou/dien}
Guorui Zhou, Na Mou, Ying Fan, Qi Pi, Weijie Bian, Chang Zhou, Xiaoqiang Zhu,
  and Kun Gai.
\newblock Deep interest evolution network for click-through rate prediction.
\newblock In {\em {AAAI}}, 2019.

\bibitem{kdd/2018/zhou/din}
Guorui Zhou, Xiaoqiang Zhu, Chengru Song, Ying Fan, Han Zhu, Xiao Ma, Yanghui
  Yan, Junqi Jin, Han Li, and Kun Gai.
\newblock Deep interest network for click-through rate prediction.
\newblock In {\em {KDD}}, 2018.

\bibitem{ijcv/2022/zhou/vpt}
Kaiyang Zhou, Jingkang Yang, Chen~Change Loy, and Ziwei Liu.
\newblock Learning to prompt for vision-language models.
\newblock {\em International Journal of Computer Vision}, 130(9):2337--2348,
  2022.

\bibitem{icml/2021/zhu/datafree}
Zhuangdi Zhu, Junyuan Hong, and Jiayu Zhou.
\newblock Data-free knowledge distillation for heterogeneous federated
  learning.
\newblock In {\em {ICML}}, 2021.

\end{thebibliography}
}
\balance

\clearpage

\appendix
\onecolumn

\section{Supplementary Notes on the Evaluation}
\label{sec:exp_sup:set}

For pretraining the original backbone model, we consistently set the batch size to 32 and the weight decay to 0.0005.
Regarding the learning rate, we have tried different recipes and chosen the one which achieves the highest accuracy on the validation set for each dataset: 
\begin{description}
\item[\inaturalist] We use a momentum factor of 0.9 and set the initial learning rate to 0.01 for the last layer and 0.001 for the other layers, and let it decay by 0.1 every 10 epochs.
\item[\celeba] We set a constant learning rate of 0.01 with a momentum factor of 0.9.
\item[\femnist and \ucf] We set a constant learning rate of 0.01 without momentum.
\end{description}

For the proposed \algname and the two baselines of fine-tuning and prompt tuning, we consistently set the batch size to 16, which is the best choice among 2, 4, 8, 16, and 32, and set the learning rate to a constant value of 0.01.
A special case is \celeba, where the number of samples per client is small, often below the default value of 16, and we make each client one batch.
Regarding the number of attention heads $H$ in \algname, we have tried the values of 1, 2, 4 and 8, and choose $H=4$ which achieves the highest accuracy on the validation dataset.
For the dimension of the prompt tokens in prompt tuning, we have tried, $D/4$, $D/2$, $D$, $2D$ and $4D$, and choose $D/2$ which achieves the highest accuracy on the validation dataset.

We conduct evaluation on machines with operating system Ubuntu 18.04.3, CUDA version 11.4, python version 3.7.13, and two NVIDIA GeForce RTX 2080Ti GPUs.
For each run, we train the model for at most 100 epochs, stop early if the accuracy on the validation dataset has not improved for 10 epochs, and adopt the model with the highest accuracy on the validation dataset. Each run takes roughly ten hours to complete. For each result that requires random weight initialization, we repeat with three random seeds and report the average accuracy on the testing dataset.

Source code and the detailed instructions to reproduce the results are available in our supplementary materials.

\end{document}